# Neural Networks for Emotion Classification

by
Yafei Sun

August 2003

# ACKNOWLEDGMENTS

I would like to express my heartfelt gratitude to Dr. Michael S. Lew for his support, invaluable guidance, time, and encouragement. I'd like to express my sincere appreciation to Dr. Nicu Sebe for his countless ideas and advice, his encouragement and sharing his valuable knowledge with me. Thanks to Ernst Lindoorn and Jelle Westbroek for their help on the experiments setup. Thanks to all the friends who participated in the construction of the authentic emotion database. I would also like to Dr. Andre Deutz, Miss Riet Derogee and Miss Rachel Van Der Waal for helping me out with a problem on the scheduling of my oral presentation of master thesis. Finally a special word of thanks to Dr. Erwin Bakker for attending my graduation exam ceremony.

# Neural Networks for Emotion Classification

## Table of Contents

# Neural Networks for Emotion Classification

## Abstract

It is argued that for the computer to be able to interact with humans, it needs to have the communication skills of humans. One of these skills is the ability to understand the emotional state of the person. This thesis describes a neural network-based approach for emotion classification. We learn a classifier that can recognize six basic emotions with an average accuracy of 77% over the Cohn-Kanade database. The novelty of this work is that instead of empirically selecting the parameters of the neural network, i.e. the learning rate, activation function parameter, momentum number, the number of nodes in one layer, etc. we developed a strategy that can automatically select comparatively better combination of these parameters. We also introduce another way to perform back propagation. Instead of using the partial differential of the error function, we use optimal algorithm; namely Powell's direction set to minimize the error function. We were also interested in construction an authentic emotion databases. This is a very important task because nowadays there is no such database available. Finally, we perform several experiments and show that our neural network approach can be successfully used for emotion recognition.

## 1  Introduction

Human beings possess and express emotions in everyday interactions with others. Emotions are often reflected on the face, in hand and body gestures, in the voice, to express our feelings or liking.

Recent psychology research has shown that the most expressive way humans display emotions is through facial expressions. Mehrabian [12] indicated that the verbal part of a message contributes only for 7% to the effect of the message as a whole, the vocal part (e.g. voice intonation) for 38%, while facial expressions contributes for 55% to the effect of the speaker's message. In addition to providing information about the affective state, facial expressions also provide information about cognitive state, such as interest, boredom, confusion, and stress, and conversational signals with information about speech emphasis and syntax.

While a precise, generally agreed definition of emotion does not exist, it is undeniable that emotions are an integral part of our existence, as one smiles to show greeting, frowns when confused, or raises one's voice when enraged. The fact that we understand emotions and know how to react to other people's expressions greatly enriches the interaction. There is a growing amount of evidence showing that emotional skills are part of what is called "intelligence". Computers today, on the other hand, are still quite "emotionally challenged." They neither recognize the user's emotions nor possess emotions of their own [1].

In order to enrich human-computer interface from point-and-click to sense–and-feel, to develop non-intrusive sensors, to develop lifelike software agents such as devices, which can express and understand emotion. Since computer systems with this capability have a wide range of applications in different research areas, including security, law enforcement, clinic, education, psychiatry and telecommunications [2], a new wave of interest in researching on emotion recognition has recently risen to improve all aspects of the interaction between humans and computers.

This emerging field has been a research interest for scientists from several different scholastic tracks, i.e. computer science, engineering, psychology, and neuroscience [1]. In the past 20 years there has been much research on recognizing emotion through facial expressions. This research was pioneered by Paul Ekman [19] who started his work from the psychology perspective. In the early 1990s the engineering community started to use these results to construct automatic methods of recognizing emotions from facial expressions in images or video [2] based on various techniques of tracking [26].

An important problem in the emotion recognition field is the lack of agreed upon benchmark database and methods for compare different methods' performance. The Cohn-Kanade database is a step in this direction [1].

There are several approaches taken in the literature for learning classifiers for emotion recognition [1] [5]:

1. **The static approach**. Here the classifier classifies each frame in the video to one of the facial expression categories based on the tracking results of that frame. Bayesian network classifiers were commonly used in this approach. While Naive-Bayes classifiers were often successful in practice, they use a very strict and often unrealistic assumption that the features are independent given the class. Therefore, another approach using Gaussian TAN classifiers have the advantage of modeling dependencies between the features without much added complexity compared to the Naive-Bayes classifiers. TAN classifiers have an additional advantage in that the dependencies between the features, modeled as a tree structure, are efficiently learned from data and the resultant tree structure is assured to maximize the likelihood function.

2. **The dynamic approach**. These classifiers take into account the temporal pattern in displaying facial expression. Hidden Markov model (HMM) based classifiers for facial expression recognition has been previously used in recent works. Cohen and Sebe [1] further advanced this line of research and proposed a multi-level HMM classifier, combining the temporal information, which allowed not only to perform the classification of a video segment to the corresponding facial expression, as in the previous works on HMM based classifiers, but also to automatically segment an arbitrary long video sequence to the different expressions segments without sorting to empirical methods of segmentation.

In this work we tried a neural network based approach using two databases: the Cohn-Kanade database and a database we constructed using the authentic data. The novelty of this work is that instead of empirically selecting the parameters of the neural network, i.e. the learning rate, activation function parameter, momentum number, the number of nodes in one layer, etc., we developed a strategy that can automatically select the best combination of these parameters. We also introduce another way to perform back propagation. Instead of using the partial differential of the error function, we use an optimal algorithm, namely, the Powell's direction set algorithm to minimize the error function. We were also interested in construction an authentic emotion databases. Because of the difficulty to get authentic data, all the experiments reported in the literature until now are done upon some databases that are constructed by telling people to show some special emotions, and thus not containing completely natural emotions.

The rest of the thesis is organized in the following way. Chapter 2 presents the current background research in emotion recognition. Chapter 3 gives a brief introduction to neural networks and the well-know back propagation model. Chapter 4 describes the optimal algorithms that are going to be used later in the experiments. Chapter 5 presents the different designs of neural networks for emotion classification. Our experiments and results are given in Chapter 6. Chapter 7 gives the discussion and the remaining open issues.

## 2   Background Research in Emotion Recognition

Since the early 1970s, Paul Ekman and his colleagues have performed extensive studies of human facial expression [20]. They found evidence to support universality in facial expressions. These "universal facial expressions" are those representing happiness, sadness, anger, fear, surprise, and disgust. They studied facial expressions in different cultures, including preliterate cultures, and found much commonality in the expression and recognition of emotions on the face. However, they observed differences in expressions as well and proposed a model in which the facial expressions are governed by "display rules" in different social contexts. For example, Japanese subjects and American subjects showed similar facial expressions while viewing the same stimulus film. However, in the presence of authorities, the Japanese viewers were more reluctant to show their real expressions. On the other hand, babies seem to exhibit a wide range of facial expressions without being taught, thus suggesting that these expressions are innate.

Ekman and Friesen [19] developed the Facial Action Coding System (FACS) to code facial expressions where movements on the face are described by a set of action units (AUs). Each AU has some related muscular basis. This system of coding facial expressions is done manually by following a set of prescribed rules. The inputs are still images of facial expressions, often at the peak of the expression. This process is very time-consuming.

Ekman's work inspired many researchers to analyze facial expressions by means of image and video processing. By tracking facial features and measuring the amount of facial movement, they attempt to categorize different facial expressions. Recent work on facial expression analysis and recognition [13] has used these "basic expressions" or a subset of them. In [4], Pantic and Rothkrantz provide an in depth review of many of the research done in automatic facial expression recognition in recent years.

The work in computer-assisted quantification of facial expressions did not start until the 1990s. Mase [3] used optical flow (OF) to recognize facial expressions. He was one of the firsts to use image-processing techniques to recognize facial expressions. Lanitis et al. [6] used a flexible shape and appearance model for image coding, person identification, pose recovery, gender recognition, and facial expression recognition. Black and Yacoob [16] used local parameterized models of image motion to recover non-rigid motion. Once recovered, these parameters were used as inputs to a rule-based classifier to recognize the six basic facial expressions. Yacoob and Davis [17] computed optical flow and used similar rules to classify the six facial expressions. Rosenblum, Yacoob, and Davis [21] also computed optical flow of regions on the face then applied a radial basis function network to classify expressions. Essa and Pentland [22] used an optical flow region-based method to recognize expressions. Donato et al. [23] tested different features for recognizing facial AUs and inferring the facial expression in the frame. Otsuka and Ohya [24] first computed optical flow, then computed the 2D Fourier transform coefficients, which were used as feature vectors for a hidden Markov model (HMM) to classify expressions. The trained system was able to recognize one of the six expressions near real-time (about 10 Hz). Furthermore, they used the tracked motions to control the facial expression of an animated Kabuki system [25]. Martinez [28] introduced an indexing approach based on the identification of frontal face images under different illumination conditions, facial expressions, and occlusions. A Bayesian approach was used to find the best match between the local observations and the learned local features model and an HMM was employed to achieve good recognition even when the new conditions did not correspond to the conditions previously encountered during the learning phase. Oliver et al. [29] used lower face tracking to extract mouth shape features and used them as inputs to an HMM based facial expression recognition system (recognizing neutral, happy, sad, and an open mouth).

These methods are similar in that they first extract some features from the images, then these features are used as inputs into a classification system, and the outcome is one of the preselected emotion categories. They differ mainly in the features extracted from the video images and in the classifiers used to distinguish between the different emotions.

## 2.1 An Ideal System for Facial Expression Recognition: Current Problems

There are three main factors to construct a Facial Expression Recognition system, namely face detection, facial feature extraction, and emotion classification. An ideal emotion analyzer should recognize the subjects regardless of gender, age, and any ethnicity. The system should be invariant to different lightening conditions and distraction as glasses, changes in hair style, facial hair, moustache, beard, etc. and also should be able to "fill in" missing parts of the face and construct a whole face. It should also perform robust facial expression analysis despite large changes in viewing condition, rigid movement, etc. A good reference system is the human visual system [4]. The current systems are far from ideal and they have a long way to achieve these goals.

## 2.2 Face Detection and Feature Extraction

Most systems detect face under controlled conditions, such as without facial hair/glasses, any rigid head movement, the first frame should be a neutral emotion etc, and thus nowadays, arbitrary face detection has drawn great intention [4].

Normally the face detection is done in 2 ways. In the holistic approach, the face is determined as a whole unit, while in an analytic approach only some important facial features are detected.

After the face is detected, there are 2 ways to extract the features. In the holistic face model, a template-based method is used. In the analytic face model, featured-based methods will be used to track the facial features while people are showing the facial expression.

In our system, we mainly focus on the emotion classification part, not on face detection or on facial feature extraction. For the extraction of the facial features we use the real time facial expression recognition system developed by Sebe and Cohen [1] (see Figure 2.1). This system is composed of a face tracking part, which outputs a vector of motion features of certain regions of the face. The features are used as inputs to a classifier.

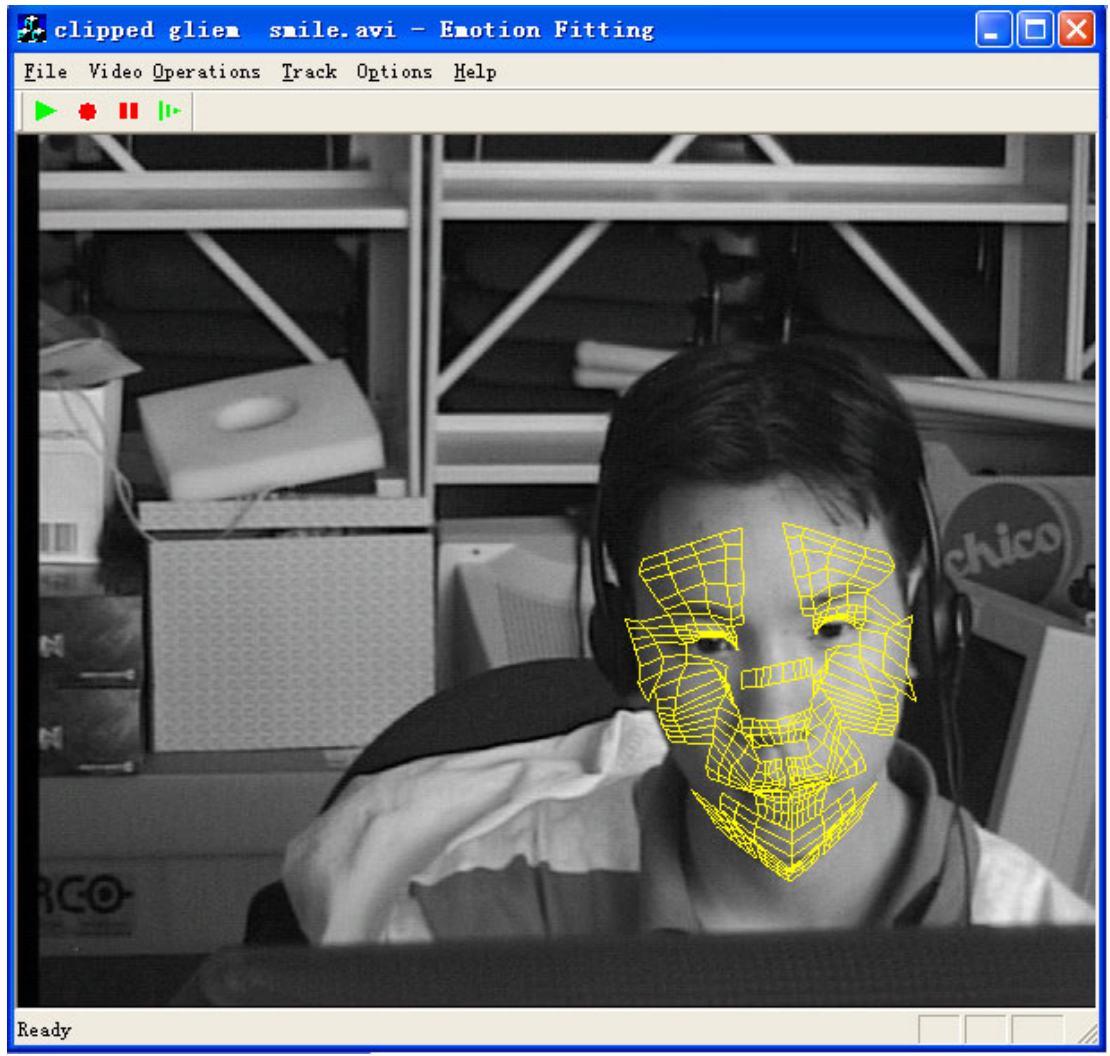

**Figure 2.1.1**

A snap shot of the real time face tracker. On the right side is a wireframe model overlaid on a face being tracked. The example is from the authentic database we created.

This face tracker uses a model-based approach where an explicit 3D-wireframe model of the face is constructed. In the first frame of the image sequence, landmark facial features such as the eye corners and mouth corners are selected interactively. The generic face model is then warped to fit the selected facial features. The face model consists of 16 surface patches embedded in Bezier volumes. The surface patches defined this way are guaranteed to be continuous and smooth. Changing the locations of the control points in the Bezier volume can change the shape of the mesh. Before describing the Bezier volume, we begin with the Bezier curve.

Given a set of control points $b_0, b_1,..., b_n$, the corresponding Bezier (or Bernstein-Bezier curve) is given by

$$x(u) = \sum_{i=0}^{n} b_i B_i^n(u) = \sum_{i=0}^{n} b_i \binom{n}{i} u^i (1-u)^{n-i}$$

where the shape of the curve is controlled by the control points $b_i$ and $u$ ranging between [0,1]. As the control points are moved, a new shape is obtained according to the Bernstein polynomials $B_n(u)$ in the above equation. The displacement of a point on the curve can be described in terms of linear combinations of displacements of the control points.

The Bezier volume is a straightforward extension of the Bezier curve and is defined by $V=BD$ written in matrix form. In this equation, $V$ is the displacement of the mesh nodes, $D$ is a matrix whose columns are the control point displacement vectors of the Bezier volume, and $B$ is the mapping in terms of Bernstein polynomials. In other words, the change in the shape of the face model can be described in terms of the deformations in $D$.

Once the model is constructed and fitted, head motion and local deformations of the facial features such as the eyebrows, eyelids, and mouth can be tracked. First the 2D image motions are measured using template matching between frames at different resolutions. Image templates from the previous frame and from the very first frame are both used for more robust tracking. The measured 2D image motions are modeled as projections of the true 3D motions onto the image plane. From the 2D motions of many points on the mesh, the 3D motion can be estimated by solving an overdetermined system of equations of the projective motions in the least squared sense. The recovered motions are represented in terms of magnitudes of some predefined motion of various facial features. Each feature motion corresponds to a simple deformation on the face, defined in terms of the Bezier volume control parameters. We refer to these motion vectors as Motion-Units (MU's). Note that they are similar but not equivalent to Ekman's AU's and are numeric in nature, representing not only the activation of a facial region, but also the direction and intensity of the motion. The MU's used in the face tracker are shown in Figure 2.2 and are described in Table 2.1.

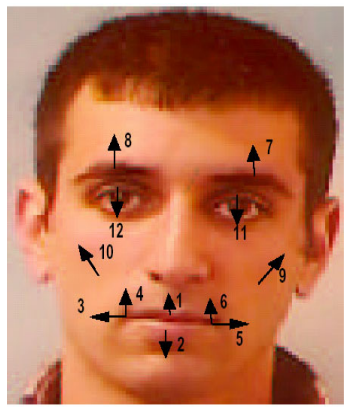

**Figure 2.2.**

The 12 facial motion measurement

| MU | Description |

| 1 | Vertical movement of the center of upper lip |
| 2 | Vertical movement of the center of lower lip |
| 3 | Horizontal movement of left mouth corner |
| 4 | Vertical movement of left mouth corner |
| 5 | Horizontal movement of right mouth corner |
| 6 | Vertical movement of right mouth corner |
| 7 | Vertical movement of right brow |
| 8 | Vertical movement of left brow |
| 9 | Lifting of right cheek |
| 10 | Lifting of left cheek |
| 11 | Blinking of right eye |
| 12 | Blinking of left eye |

**Table 2.1**

Motion Units used in the face tracker

Each facial expression is modeled as a linear combination of the MU's:

$$V = B[D_0 D_1 \ldots D_m] \begin{bmatrix} P_0 \\ P_1 \\ \vdots \\ P_m \end{bmatrix} = BDP$$

where each of the $D$ corresponds to a MU, and the $P_i$ are the corresponding magnitudes (or coefficients) of each deformation. The overall motion of the head and face is:

$$R(V_0 + BDP) + T$$

where *R* is the 3D-rotation matrix, *T* is the 3D-translation matrix, and $V_0$ is the initial face model. The MU's are used as the basic features for the classification scheme [1].

## 3  Neural Network Background

One of the main purposes of artificial intelligence is to simulate the human brain. Nowadays, the storage capability, computation speed, and other logical function of numeric computer already exceed human brain, but still there are a lot of human brain's functions that cannot be simulated by a computer, i.e. human being's recognition and decision capability. The neural network theory is based on the research of simulating human being's information processing methods.

### 3.1  Conformation and function of the neuron

Artificial neural networks were originally designed to model in some way the functionality of the biological neural networks, which are a part of the human brain. Our brains contain about $10^{11}$ neurons. Each biological neuron consists of a cell body, a collection of dendrites, which bring electrochemical information into the cell, and an axon, which transmits electrochemical information out of the cell.

A neuron produces an output along its axon i.e. *it fires* when the collective effect of its inputs reaches a certain threshold. The axon from one neuron can influence the dendrites of another neuron across junctions called synapses. Some synapses will generate a positive effect in the dendrite, i.e. one which encourages its neuron to fire, and others will produce a negative effect, i.e. one, which discourages the neuron from firing. A single neuron receives inputs from about $10^5$ synapses and the total number of synapses in our brains may be of the order of $10^{16}$. It is still not clear exactly how our brains learn and remember but it appears to be associated with the interconnections between the neurons (i.e. at the synapses).

Artificial neural networks try to model this low-level functionality of the brain. This contrasts with high level symbolic reasoning in artificial intelligence, which tries to model the high level reasoning processes of the brain. When we think we are conscious of manipulating concepts to which we attach names (or symbols) e.g. for people or objects we are not conscious of the low level electrochemical processes which are going on underneath. The argument for the neural net approach to AI is that, if we can model the low-level activities correctly, the high level functionality may be produced as *an emergent property*.

## 3.2   A Short History of Artificial Neural Networks

- In 1943, the psychologist McCulloch and the mathematician Pitts created the first neural network model called the MP model.
- 1943 – 1970, a variety of models and learning algorithm were developed: Hebb came out with his learning rule; the perceptron model was developed by Rosenblatt, etc.
- 1970 - 1980, it was the low-tide period for this area; the theory developed until this time had a lot of incompleteness.
- 1980 - 1990, after the American physical scientist J. Hopfield promoted the Hopfield neural network (HNNS), and made progress on the problem of traveling sales man, a number models and algorithms starting to emerge.

- After 1990, this area is in a constant progress. Practically, there are a lot of applications in many areas including pattern recognition, data compression, coding, cipher analysis, stock analysis, economic management, and optimal control.

## 3.3 Selection of neural network model and the theory of back propagation

In the articles from the literature we surveyed, back propagation [4, 14] was the commonly used model for emotion classification. Is there any special reason that makes back propagation better than the numerous other neural network models? Or should we try some new models to see if they work better than the "popular" back propagation? Some answers to these questions can be found in [15] (see Table 3.3.1). SOM, known also as Kohonen's self-organizing map, which is a clustering technique that can be used to automatically provide insight into the nature of data without supervision. We can transform this unsupervised neural network into a supervised LVQ (Learning Vector Quantum) neural network, which is another clustering technique. In the table, we can infer that when handling noise and multiple input of data, back propagation performs better than SOM. LVQ is excellent for classification, but when handling noise is a little bit worse than back propagation. Taking all these facts into consideration, we decided to use back-propagation as our main method.

| Network Type | Problem Type | | Data Constraints | |
|---|---|---|---|---|
| | Classification | Prediction | Noisy: Ability to cope with very noisy data (such as in stock market models, marketing, etc) | Many inputs: Effectiveness with a large number of input fields |
| Back-Propagation | Good | Excellent | Excellent | Excellent |
| SOM | Good | Average | Good | Good |
| LVQ | Excellent | Very poor | Good | Excellent |

Table 3.3.1

A comparison of different neural network models. SOM refers to Kohonen's self-organizing map and LVQ refers to Learning Vector Quantum neural network.

Figure 3.3.1 shows a simple neuron (perceptron) in a neural network with *n-1* inputs, 1 bias, and 1 output. The total input stimuli to this neuron in the output layer is:

$$z_{in} = \sum_{i=0}^{n} x_i w_i = x_0 w_0 + x_1 w_1 + x_2 w_2 + \ldots + x_n w_n$$

The output of this neuron is the activation function *f($z_{in}$)*, where f is activation function. Through out the experiments, activation function is defined as:

$$f(x) = \frac{1}{(1 + e^{-\sigma x})}$$

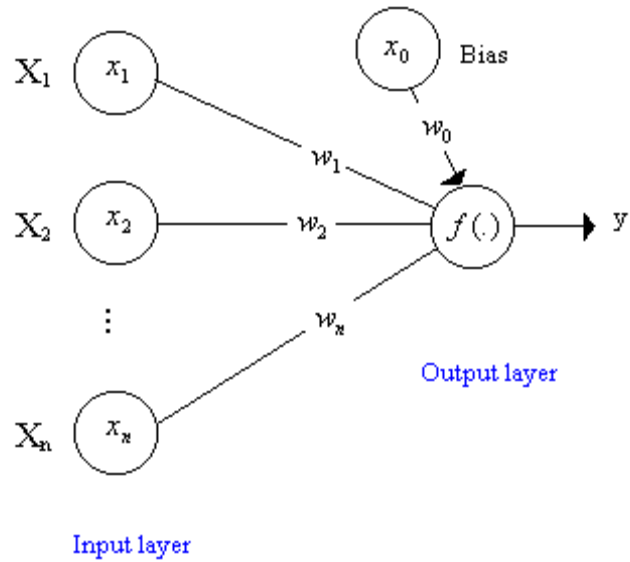

**Figure 3.3.1**

A single perceptron in a neural network

After the neuron outputs its value, generally there is a difference between this value and its correct value, the target value. The difference is defined as the error function:

$$E = \frac{1}{2}\sum_{k=1}^{m}(t_k - y_k)^2$$

We use the partial differential of this error function to modify the weights backwards, thus make the error decrease. This procedure is called **back propagation**. The modification equation of weights using the partial differential of the error function is given by:

$$\Delta w_{jk} = -\alpha \frac{\partial E}{\partial w_{jk}}$$

where α is the learning rate, and $w_{jk}$ is the weight between neuron $j$ and neuron $k$.

After a number of modifications of weights by the error function's partial differential, the error will decrease to some extent that can be accepted. At this point, the neural network can be used to test data. Figure 3.3.2 is the complete figure of one hidden layer network with multiple neurons, where x is input, z represents hidden nodes, y represents output, t is the target value, w and v represents weights, bias is usually set to 1.

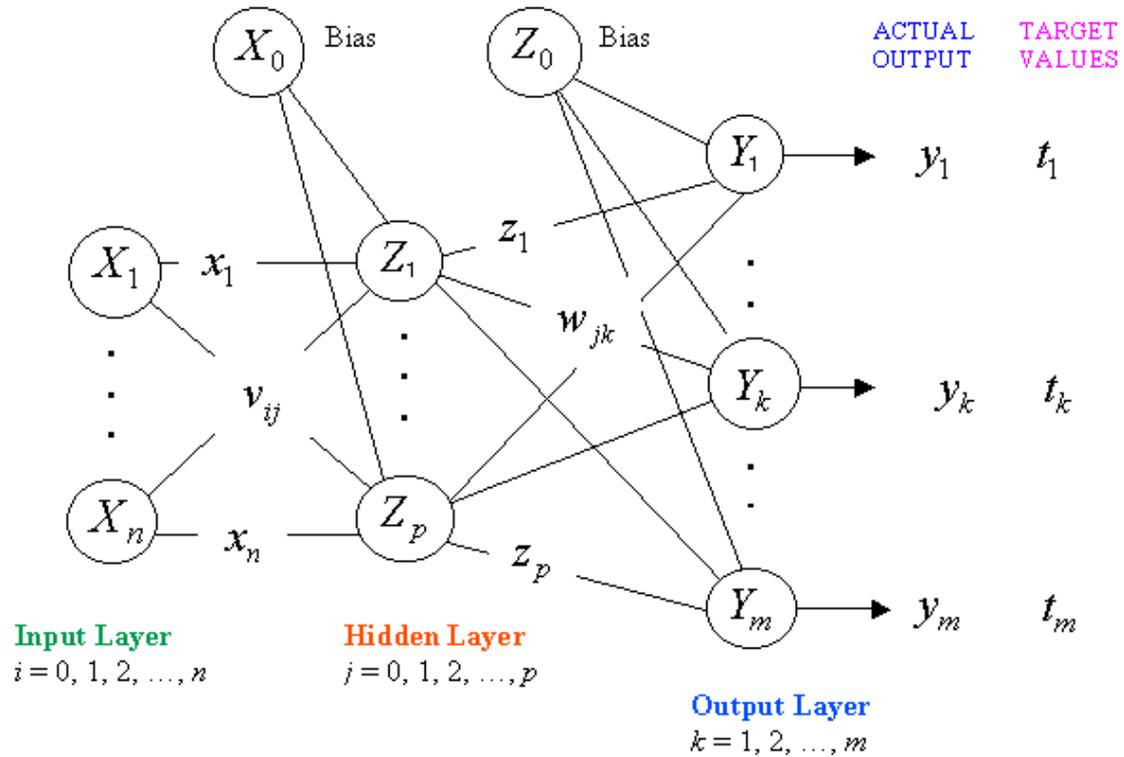

**Figure 3.3.2**

The complete figure of one hidden layer network with multiple neurons

The algorithm for one hidden layer is given on the next page. In this algorithm, *i* represents the number of input nodes, *j* is number of hidden nodes, *k* is number of output nodes, $w_{jk}$ represents the weights between neuron *k* in output layer and neuron *j* in hidden layer, and $v_{ij}$ represents the weights between neuron *j* in hidden layer and neuron *i* in input layer.

Theoretically, two-hidden-layer back propagation neural network can simulate any arbitrary function, so we also tried the two-hidden-layer network to see if it works better than one-hidden-layer network. For two-hidden-layer network, the step action is similar to that of one hidden layer, except adding and modifying the weights of one extra layer.

## Step action of one-hidden-layer back propagation

Initialize weights to small random values

while (stopping condition false)

   {

      for ( each training pattern)

         {

            Set all $x_i$ for i =1,2,...,n

            for *j= 1,2, ..., p*

              {

$$zin_j = \sum_{j=0}^{n} x_i v_{ij}$$

$$z_j = f(zin_j)$$

              }

            for *k= 1,2, ..., m*

              {

$$yin_k = \sum_{j=0}^{p} z_j w_{jk}$$

$$y_k = f(yin_k)$$

              }

$$\delta_k = (t_k - k_k) f'(yin_k)$$
$$\Delta w_{jk} = \alpha \delta_k z_j$$
$$\delta_j = f'(zin_j) \sum_{k=1}^{m} \delta_k w_{jk}$$
$$\Delta v_{ij} = \alpha \delta_j x_j$$
$$w_{jk}(new) = w_{jk}(old) + \Delta w_{jk}$$
$$v_{ij}(new) = v_{ij}(old) + \Delta v_{ij}$$

}/* end of for loop */

}/* end of while loop */

# 4  Optimal methods: Powell and Downhill Simplex

The essence of the training procedure of a neural network is to find a set of weights that can minimize the error function. The typical neural network way is to propagate backwards the error function's partial differential. We are interested to know if the traditional optimal algorithm for neural networks works. Among the surveyed minimization and maximization methods presented in [8], there are three methods for multi-dimensional input and which do not require the function to have a first derivative. These methods are simulated annealing, downhill simplex, and Powell's method. Since there is already report on combining simulated annealing with neural nets, we decided to try the last two methods in our experiments.

## 4.1  Powell's Direction Set Method

Assume we start at a point $P$ in an $N$-dimensional space, and proceed from there along a direction defined by a vector $E$. Then, any function of $N$ variables $f(P)$ can be minimized in the following way:

> Take the unit vectors $e_1, e_2, ..., e_N$ as a set of directions, move along the first direction to its minimum, then from there along the second direction to its minimum, and so on, cycling through the whole set of directions as many times as necessary, until the function stops decreasing. [8]

Initialize the set of directions $u_i$ to the basis vectors (the already defined vectors or very small random value), $u_i = e_i$ with $i = 1,...,N$.
Now repeat the following sequence of steps until objective function stops decreasing:

Save the starting position as $P_0$.

For $i = 1,\ldots,N$, move $P_{i-1}$ to the minimum along direction $u_i$ and call this point $P_i$.

For $i = 1,\ldots,N-1$, set $u_i = u_{i+1}$.

Set $u_N = P_N - P_0$.

Move $P_N$ to the minimum along direction $u_N$ and call this point $P_0$ [8].

## 4.2  Downhill Simplex Algorithm

A simplex is the geometrical figure in $N$ dimensions, of $N + 1$ points (or vertices). In two dimensions, a simplex is a triangle. In three dimensions it is a tetrahedron [8]. The downhill simplex method starts not just with a single point, but with $N + 1$ points, defining an initial simplex. Assume any one of these points as being the initial starting point $P_0$, then take the other $N$ points to be $P_i = P_0 + e_i$ where the $e_i$'s are $N$ unit vectors [8]

The downhill simplex method now takes a series of steps, most steps just moving the point of the simplex where the function is largest ('highest point') through the opposite face of the simplex to a lower point. These steps are called reflections, and they are constructed to conserve the volume of the simplex (hence maintain its non-degeneracy). When it can do so, the method expands the simplex in one or another direction to take larger steps. When it reaches a 'valley floor,' the method contracts itself in the transverse direction and tries to ooze down the valley. If there is a situation where the simplex is trying to 'pass through the eye of a needle,' it contracts itself in all directions, pulling itself in around its lowest (best) point. It is then possible to terminate when the vector distance moved in that step is fractionally smaller in magnitude than some tolerance [8].

In following chapter, there is detailed description about how to combine the presented algorithms into the neural network.

# 5 Design of Neural Networks for Emotion Recognition

## 5.1 General description

Now with all the background knowledge, we can start the design of neural networks for emotion recognition. The 12 features data obtained from the face tracker are used as the input of the 12 input nodes in a neural network. The output layer contains 2-7 nodes that represent the emotion categories, depending on different networks. There are 1 or 2 hidden layers and the number of hidden nodes ranges from 1 to 29x29. The learning rate, momentum number, and the parameter of the sigmoid activation function are automatically adjusted during the training procedure. In some networks, the Powell's method is considered, while in others, a set of empirical ways are combined, i.e. take the peak frames of the emotion data sequence, sort the training set, delete some of the emotions, normalize the output, set threshold to the weights, etc. The test results are based on Cohn-Kanade database and on the authentic database separately. The activation function we used is the sigmoid function. In the following sections, we present a description of all parameters, and their combinations' results in experiments.

## 5.2 Weights

Back-propagation is a gradient descent search, so it is easy to stop at a local minimum, while randomly selected weights help to avoid this [14]. If the weights are too large, the networks tend to get saturated. The solution is to ensure that, after weight initialization and before learning, the output of all the neurons is small value between [-0.5, 0.5]. We initialize the weights by a random function and ignore those weights that are larger than a specific threshold which can also be adjusted as one of the parameter of the network.

One question is if during the training procedure, should we constrain the weights as well? This partially depends on how large the input is, because the sigmoid function is very close to one when the input is greater than 10 [14]. Since our feature data for input is very small, usually smaller than 2, we set the threshold of the weights, ignoring those weights that exceed this constraint during training. It came out that this did not make much difference at improving the hit rate. On the other hand, when we tried a very strict threshold in a 2-hidden-layer neural network during the training procedure, sometimes it led awful performance of the 2-layer network. This is because the parameters of the activation we set in a 2-hidden-layer network were not proper for that threshold, and caused the saturation of the neuron. Hence, we gave a very large threshold after initialization to the weights to avoid similar problems.

Another thing we should take care of is the starting point, which can also affect the search direction to find a good local minimum (see Figure 5.2.1). If we start at point A, we obtain the global minimum, while from C, we get the local minimum. So we should try different staring points by initializing the weights with different random values. We tested this in some of the networks and found that the accuracy curve fluctuates, but not too much. For details, see Figure 6.4.1 and Figure 6.4.2 in Chapter 6.

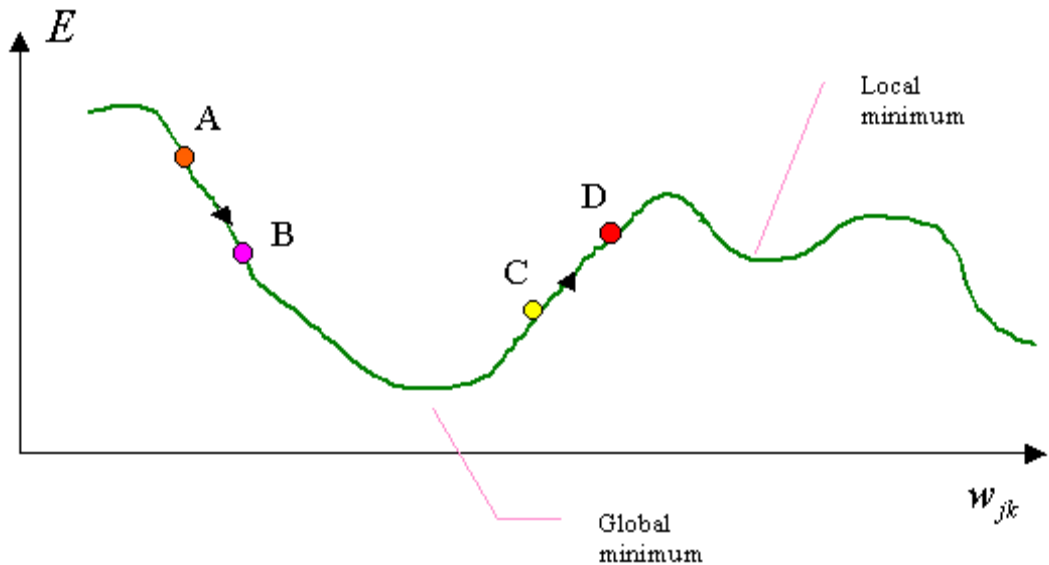

**Figure 5.2.1**

Local minimum and global minimum

## 5.3  Key parameters and their combination

In design of neural networks, there are some critical parameters that need to be set, i.e. the learning rate $\alpha$, the momentum number $\lambda$, and the activation function parameter $\sigma$.

The speed of the learning is governed by the learning rate $\alpha$. The momentum number carries along the weight change, thus it tends to smooth the error-weight space. An improper value of $\sigma$ will cause the neuron saturated.

In general, the performance of neural networks will be very awful, if these values are not chosen correctly. Unfortunately, there are no precise rules or mathematics definition upon when one should choose what particular value of these parameters. Normally, the setting of the parameters is done empirically. Does it help in finding better combinations if we let the computer do part of the job? We tried this in the following way.

First, we defined three different categories. The increase or decrease step size of α, λ, and σ is given by input or macro definition. This depends on how frequently these categories should be changed during training procedure, e.g., for those categories that need little interference, we give a macro definition for training efficiency. When better accuracy occurs, the rates together with the parameters which lead to this accuracy are recorded in a file. We repeat the training until the accuracy stops improving for some turns. When testing, we construct a neural network by reading these parameters from the file.

Since the training with the complete combinations of these parameters costs quite a long time, we only tried a part of these three parameters' combinations. Therefore, it is possible to miss some better combinations of them.

## 5.4 Number of hidden layers and hidden nodes

The number of hidden layers and hidden nodes will also affect the performance. The network will not produce a good model of the problem if too few hidden neurons are used, while too many hidden neurons will lead to poor generalization. Only with a proper number of hidden neurons, right learning rate, momentum number, and activation function parameter, the neural network can work well. Again, it is very time consuming to automatically try all the different number of hidden nodes with different other key parameters' combinations. Thus, although the code was designed to automatically select the combination, in practical training, we still empirically select some typical sizes (number of hidden nodes) of network as the starting point. For the results, see the tables in chapter 6.3.

## 5.5   Combine the Powell's direction set

We tried a lot of parameter combinations of back propagation network, but it seems the recognition rate stays at around 60%-76.8% (results on Cohn-Kanade database). This fact urged us to try some improvements. One obvious choice is to try some of the existing optimal algorithms. They regard the neural network as a function, which returns an error that can be minimized.

We surveyed several minimization algorithms [8]. Only simulated annealing, Powell's method, and downhill simplex can handle multi-dimensional inputs, and do not need the function's first derivative. We first tried the downhill simplex, but when trying this method on normal function such as

$$f(x, z)=(x-20)*(x-20)+(z-400)*(z-400),$$

it could not handle more than 2 dimensional input, so we stopped at this point.

As for Powell's method, we tried up to 6 dimensional input function:

$$Y(x_1, x_2, x_3, x_4, x_5, x_6)=(x_1-20.0)*(x_1-20.0)+(x_2-40.0)*(x_2-40.0)+(x_3-3.0)$$

$$*(x_3-3.0)+(x_4-40.0)*(x_4-40.0)++(x_5-30)*(x_5-30)$$

$$+(x_6-95)*(x_6-95)$$

It can successfully find all the values that can minimize y: 20, 40, 3, 40, 30, 95. So we assume it can work with arbitrary dimension.

We set the weights of the network to be the input $X$ of the function, the dimension is the number of all the weights. If it is a 2_hidden_layer network, just calculate the number of all the weights' no matter which layer the weight is at. Pick up a random direction set with very small value, then start error minimization and modify the weights by Powell's procedure.

Unfortunately, this time we failed to obtain a good recognition rate. Comparing with normal back propagation neural network, this method took several minutes to reach the rate of 30%, while back propagation neural network needed no more than half minute to achieve the same success. So we did not continue in this direction.

A legitimate question is: does it help if we start with a trained weight set from a back propagation way? We tried this procedure, but the results oscillated so we could not draw any conclusion on whether this method will always improve with the existing trained weights.

From the performance and training speed to achieve the same recognition rate, we can conclude that neural networks with weights optimized by back propagation of error differential outperform the other methods. The reason might be that when modifying the weights by back propagation, it considers the partial differential of error function to be its guide, while in Powell's method, the directions are chosen randomly. Therefore, there is no guide information anymore, and it is easier to get trapped into a local minimum.

## 5.6 Empirical methods

### 5.6.1 Normalization of Output

Back propagation is a supervised way for training and testing. In our case, both the test data and the training data are labeled with integer ranging from 1 to 7 representing 7 possible expressions: neutral, joy, surprise, angry, disgust, fear, and sad. On the other hand, the output of the neural network in each layer is a float, so we normalized the values of the final output layer and set the max value to 1, others to 0. In some cases of our experiments, especially in the 1_hidden_layer neural networks, this procedure helped to improve the recognition rate (see figure 6.4.1 and figure 6.4.2).

### 5.6.2 Use Only Peak Frames of the Data

The frames of data are continuous, neutral emotion is always at the beginning of each emotion sequence, after which follows the series of frames of "real" emotion data. Thus in some of our networks, we picked up only the first 3 frames of the neutral data, and 3 frames of the "real" emotional data in the middle of the expression, and ignored all the other frames that mainly consist of transition emotion data. This procedure should improve the accuracy by some extent, but the drawback is that after filtering out all intermediate frames, the number of frames for training decreases sharply. This seems to be the reason for the decrease in the classification rate (see the tables in chapter 6).

### 5.6.3 Median filter

In the Cohn-Kanade data set and the authentic data set we created, a particular emotion is represented by a set of successive frames. This fact made us to try a median filtering procedure on the outputs of the neural networks with the intention of improving the classification rate.

The idea is that when a recognized emotion value (the value given by neural net output layer) is different from its next neighbor and its previous neighbor, it must be a wrong value. The median filter will reset it to the value of its neighbors. The size of the neighborhood will tune the improvement given by this method. Currently, the median filter contributes to an improvement is accuracy of about 1% to 5%, but normally.

### 5.6.4  Ignore all the neutral values

In the subset of Cohn-Kanade we use, the neutral data constitute about 1/3 of the whole training set, which might lead the neural network parameters to be tuned more by the neutral data. While in practical application the neutral data might be the least interesting expression, in some experiments we decided to ignore all the neutral data to improve the accuracy. This procedure improved the rate by some extend. For the results, see tables in Chapter 6.

### 5.6.5  Sort and select the training set

In some experiments, we sorted the training data set and used exactly the same amount of each emotion data in some of our one-hidden-layer networks, in order to balance the effect of each emotion to the whole network. The reorganization (sorting and selecting) of the data helps to improve the accuracy from 1% to10%. But how it affects the 2-hidden-layer networks, we didn't try. See the results in chapter 6.

### 5.6.6 About over-training

Over-training might occur when the number of training data over a particular value. This value will differ when the parameters change. It is quite time consuming to train each network to see how many data should be under each group of parameters, especially when we are still searching for the best parameter set. We tested the over-training in a few cases, but we didn't reach a conclusion regarding the optimal number of training data that will not cause the over training of the neural network.

# 6 Experiments

## 6.1 Experiment setup and results for authentic data

In the papers we surveyed, all the experimental results are based on databases that did not capture natural emotions, e.g. the subject was instructed to show a particular emotion in front of the camera.

An ideal way to get a natural emotion data is using a secret camera, and thus not telling the person in advance that he is being filmed. However, this type of experiments may cause some ethical problems, so we only tested with our friends and colleagues. They were told that we are a doing a psychology test or a test with the image quality after decompression. We let them watch a short period (9 minutes) of video clips extracted from an horror movie and comedy, disgusting, surprising video clips. We also showed a whole horror movie to some of them. Afterwards, when we explained them what our intention was and most of them were amused and a little bit surprised.

To find enough persons to spend 2 hours watching a whole movie in a special situation and tell them not to speak or eat, sit at a proper distance and pose correctly to allow the camera to catch their faces is a very tedious and difficult task. Therefore, we decided to use the short video clips in most cases, even if they are not long enough to build up most of the emotions, such as sad, fear, anger or even surprise and disgust. It was also difficult and time consuming to find suitable videos and movies that could arouse people's emotions immediately, i.e., it is tricky to disgust people enough to show the expression without annoying them. Before recording the real experiments, we tried a lot of video clips on some of the faculty in LIACS, and changed several times the candidate video clips. Another problem was where to hide the normal size camera for secret filming. Sometimes we used the curtain, sometimes the camera was behind a hole in a book folder, and sometimes it was in a big monkey-toy.

Despite all these difficulties, in total we tested 28 people. However, some people did not always sit in a proper position, or they were talking and moving. Others didn't even show any emotion when watching the short clips for they were only very seriously looking at the quality of the images as they were told! In the end, we selected the data from 15 persons, including 4 Chinese female, 1 Chinese male, 2 Dutch female, and 8 Dutch male. We did not put any constraint on their age.

We did this test mostly with short clips, thus it was more likely to get only some particular expressions such as neutral, joy, and sometimes disgust. The Chinese girls participating in the whole 2 hours' experiment showed several emotions such as joy, fear, disgust, and surprise. Some of them confessed they were very tensioned and scared but they didn't show the obvious emotions as we expected. They only looked very serious, while others showed surprise but blended with fear. Taking all these into account, we decided to select the data of neutral, joy, and disgust.

Figure 6.1.1 shows three frames of one person from our authentic database showing different expressions: neutral, joy and disgust. Figure 6.1.2 shows examples of two people from the Cohn-Kanade database showing 6 different expressions.

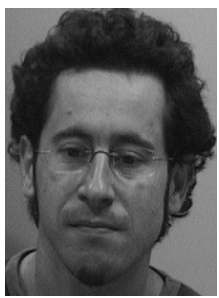 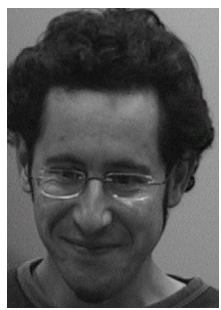 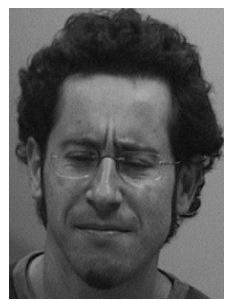

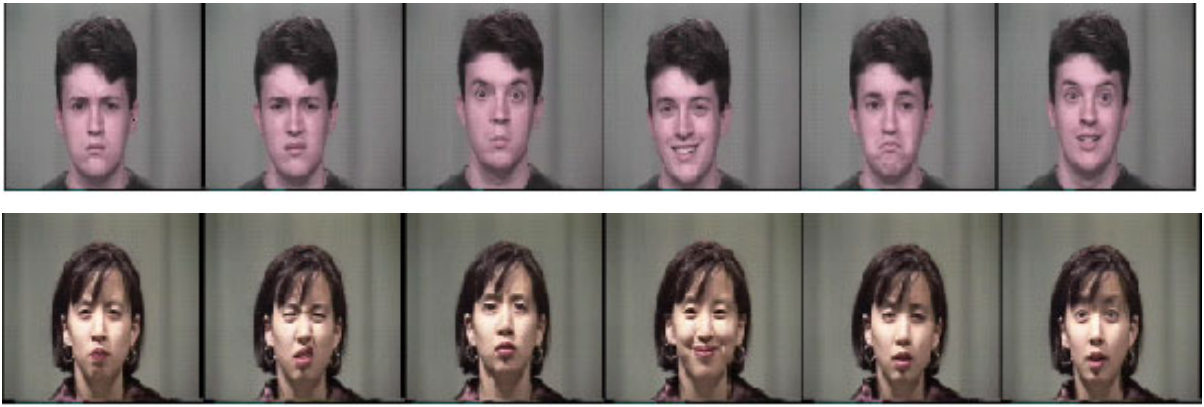

Neutral　　　　　　　　　Joy　　　　　　　　　Disgust

**Figure 6.1.1**

Images from the authentic database

Anger　　Disgust　　Fear　　Joy　　Sad　　Surprise

**Figure 6.1.2**

Images from Cohn-Kanade database

After recording the emotion, we used Adobe Premiere and Virtual Dub to annotate data, and the face tracker mentioned in Chapter 2 to extract features. In total, 11 Persons are used for the training set and the other 4 as test persons. We also did an experiment using the training set consisting of 70% of the data from 11 people, while the test set was the remaining 30% of the data. For the result, see table 6.1.1.

|  | Neutral | Joy | Disgust |
| --- | --- | --- | --- |
| Neutral | 96.08% | 3.92% | 0.0% |
| Joy | 4.67% | 95.33% | 0.0% |
| Disgust | 1.59% | 0.0% | 98.41% |

**Table 6.1.1**

Results on authentic data set. The average classification rate is 96.6%. The training set consists of 70% of the data from 11 people and the test set is the remaining 30% data.

Network parameters: 1 hidden layer, 10 hidden nodes, $\lambda= 0.55$, $\alpha= 0.42$, $\sigma= 0.3$, number of test data is 264, number of training data is 677, with median filter and normalized output.

For comparison reasons, we also tried only classifying the neutral, disgust, and joy with Cohn-Kanade. The results are given in Table 6.1.2.

|  | Neutral | Joy | Disgust |
|---|---|---|---|
| Neutral | 82.78 | 8.71 | 8.51 |
| Joy | 19.67 | 77.05 | 3.28 |
| Disgust | 25.00 | 6.82 | 68.18 |

**Table 6.1.2**

Results on Cohn-Kanade data set.

Network parameters: 1 hidden layer, 10 hidden nodes, $\lambda= 0.5$, $\alpha=0.3$, $\sigma=0.3$, test vector number 687, training vector number 2300, use median filter and normalized output. The average here is 76%, lower than that from authentic database.

Although there are 7 emotions in Cohn-Kanade database, while ours only has 3, but when only processing 3 emotions in Cohn_Kanade database, the average recognition rate is also much lower than which obtained from authentic database, see results given in table 6.1.1 and table 6.1.2.

However, it is still too early to conclude for the difference between these two results given in table 6.1.1 and table 6.1.2. Because when extracting the data for the authentic database, we picked up the peak frames manually, thus ignored the transition emotion. On the other hand, taking the peak frames from Cohn_Kanade database was done by code, with fixed peak frame position which was hard to say it was the real peak frames of that emotion.

The reason we chose the most obvious emotion data from our videos is that we want to know how well the neural networks will work with the emotion that can easily be distinguished by human. Table 6.1.1 shows it can work quite well in this case, but how it works with transition emotion data, it is unknown.

|  | Neutral | Joy | Disgust |
|---|---|---|---|
| Neutral | 100% | 0.00% | 0.00% |
| Joy | - | - | - |
| Disgust | 0.00% | 0.00% | 100% |

**Table 6.1.3**

Results on authentic data set. Test on subject 1 (not included in the training set).

Network parameters: 1 hidden layer with 10 hidden nodes, λ= 0.55, α=0.36, σ= 0.3, number of test data 22, number of training data 677, with median filter and normalized output. "-" means that the corresponding measurements were not available.

|  | Neutral | Joy | Disgust |
|---|---|---|---|
| Neutral | 100% | 0.00% | 0.00% |
| Joy | 33.33% | 66.67% | 0.00% |
| Disgust | 0.00% | 0.00% | 100% |

**Table 6.1.4**

Results on authentic data set. Test on subject 2 (not included in the training set).

Network parameters: 1 hidden layer with 10 hidden nodes, λ= 0.55, α = 0.3, σ=0.3, number of test data 100, number of training data 677, with median filter and normalized output.

|  | Neutral | Joy | Disgust |
|---|---|---|---|
| Neutral | 100% | 0 | 0 |
| Joy | 0 | 100% | 0 |
| Disgust | - | - | - |

**Table 6.1.5**

Results on authentic data set, test on subject 3 (not included in the training set).

Network parameters: 1 hidden layer with 10 hidden nodes, $\lambda = 0.55$, $\alpha = 0.36$, $\sigma = 0.3$, number of test data 83, number of training data 677, with median filter and normalized output.

|         | Neutral | Joy    | Disgust |
|---------|---------|--------|---------|
| Neutral | 100%    | 0      | 0       |
| Joy     | 0       | 93.75% | 6.25%   |
| Disgust | -       | -      | -       |

**Table 6.1.6**

Results on authentic data set. Test on subject 4 (not included in the training set).

Network parameters: 1 hidden layer with 10 hidden nodes, $\lambda = 0.55$, $\alpha = 0.3$, $\sigma = 0.3$, number of test data 17, number of training data 677, with median filter and normalized output.

From table 6.1.3, table 6.1.5 and table 6.1.6, we can conclude that even if testing subjects were not included in the train set, the recognition rate was relatively high. The reason might either be that we picked up the peak frames manually when annotating the video clips, thus ignored the transition emotion; Or it is easier to recognize authentic emotion than not natural ones.

In table 6.1.4, joy has 33% confused with neutral. We noticed that this subject was very cheerful all the time in the experiment, even the neutral emotion looked a little bit joyful, but it is not precise to say if this is the reason for the much lower recognition rate.

If we can test more persons, include the transition emotion data and obtain more emotions, it is clearer to see how neural network will work with authentic data.

## 6.2 Results of test on Cohn-Kanade data set - Empirical methods

### 6.2.1 One_hidden_layer neural networks for testing the empirical methods

6.2.1.1 The following group is result of one_hidden_layer neural networks with 40 hidden nodes, **exclude neutral, not take peak**, no momentum number, and combine with either using median filter and sorting the train data or not.

| Emotion | Joy | Surprise | Angry | Disgust | Fear | Sad |
|---|---|---|---|---|---|---|
| Joy | 77. 87 | 0. 00 | 10. 66 | 1. 64 | 9. 84 | 0. 00 |
| Surprise | 0. 00 | 92. 93 | 0. 00 | 0. 00 | 0. 00 | 7. 07 |
| Angry | 0. 00 | 0. 00 | 70. 75 | 27. 89 | 0. 00 | 1. 36 |
| Disgust | 4. 55 | 0. 00 | 3. 41 | 81. 82 | 1. 14 | 9. 09 |
| Fear | 26. 67 | 0. 00 | 4. 76 | 0. 95 | 67. 62 | 0. 00 |
| Sad | 12. 70 | 3. 97 | 8. 73 | 8. 73 | 9. 52 | 56. 35 |
| average | 74. 55 | | | | | |

**Table 6.2.1.1.1**

The result is obtained by neural network with 1 hidden layer, 40 hidden nodes, no momentum number, σ = 0.91, α = 0.81 and ignored the neutral emotion, sorted the train data set, *with median filter* which size was 3. The test set had 687 vectors and the number of original training set was 2300, after *sorting train data,* the number of train data decreased to 1446.

| Emotion | Joy | Surprise | Angry | Disgust | Fear | Sad |
|---|---|---|---|---|---|---|
| Joy | 77. 05 | 0. 00 | 10. 66 | 2. 46 | 9. 84 | 0. 00 |
| Surprise | 0. 00 | 90. 91 | 0. 00 | 0. 00 | 2. 02 | 7. 07 |
| Angry | 0. 68 | 0. 00 | 70. 75 | 27. 21 | 0. 00 | 1. 36 |
| Disgust | 5. 68 | 0. 00 | 5. 68 | 78. 41 | 1. 14 | 9. 09 |

| Emotion | | | | | |
|---|---|---|---|---|---|
| Fear | 26. 67 | 0. 00 | 3. 81 | 0. 95 | 67. 62 | 0. 95 |
| Sad | 12. 70 | 4. 76 | 8. 73 | 8. 73 | 9. 52 | 55. 56 |
| average | 73. 38 | | | | | |

**Table 6.2.1.1.2.**

The result is obtained by neural network with 1 hidden layer, 40 hidden nodes, no momentum number, σ = 0.91, α = 0.81 when ignored the neutral emotion and sorted the train data set and no *median filter*. The test had 687 vectors and train set had 2300, after *sorting train set* number of train vectors decreased to 1446.

The only difference between table 6.2.1.1.2 and table 6.2.1.1.1 is that the later one used median filter, thus caused the average rate 1.17% higher than previous one. The compare between table 6.2.1.3.1 and table 6.2.1.3.2 also showed that the medium filter will improve the rate by 0.5% -2%.

| Emotion | Joy | Surprise | Angry | Disgust | Fear | Sad |
|---|---|---|---|---|---|---|
| Joy | 72. 95 | 0. 00 | 10. 66 | 4. 92 | 11. 48 | 0. 00 |
| Surprise | 0. 00 | 92. 93 | 0. 00 | 0. 00 | 0. 00 | 7. 07 |
| Angry | 0. 00 | 0. 00 | 66. 67 | 31. 97 | 0. 00 | 1. 36 |
| Disgust | 4. 55 | 0. 00 | 2. 27 | 84. 09 | 1. 14 | 7. 95 |
| Fear | 26. 67 | 0. 00 | 4. 76 | 0. 95 | 67. 62 | 0. 00 |
| Sad | 9. 52 | 8. 73 | 8. 73 | 11. 90 | 6. 35 | 54. 76 |
| average | 73. 17 | | | | | |

**Table 6.2.1.1.3**

The result is obtained by neural network with 1 hidden layer, 40 hidden nodes, no momentum number, σ = 0.81, α = 0.71 when ignoring the neutral emotion, sorting the train data set, using median filter with size equaled 3.

Number of test vectors was 687, number of original training set was 2500, after sorting, number of train set data was 1476.

The difference between table 6.2.1.1.3 and table 6.2.1.1.1 is that the later one had 2300 train vectors while the previous had 2500, the rate decreased while neural net had 200 more vectors. This can be explained as over-training. One should note that for different neural networks in this thesis, the number of train set which causes over training may varies with different combinations of key parameters.

| Emotion | Joy | Surprise | Angry | Disgust | Fear | Sad |
|---|---|---|---|---|---|---|
| Joy | 77. 05 | 0. 00 | 2. 46 | 4. 10 | 15. 57 | 0. 82 |
| Surprise | 0. 00 | 79. 80 | 0. 00 | 0. 00 | 9. 09 | 11. 11 |
| Angry | 5. 44 | 0. 00 | 53. 06 | 38. 78 | 0. 00 | 2. 72 |
| Disgust | 5. 68 | 0. 00 | 3. 41 | 72. 73 | 4. 55 | 13. 64 |
| Fear | 37. 14 | 0. 00 | 0. 00 | 3. 81 | 56. 19 | 2. 86 |
| Sad | 8. 73 | 4. 76 | 3. 97 | 6. 35 | 7. 94 | 68. 25 |
| average | 67. 85 | | | | | |

**Table 6.2.1.1.4**

The result is obtained by neural network with 1 hidden layer, 40 hidden nodes, σ = 0.91, α = 0.71 when ignoring the neutral emotion and not *sorting the train data set*.

The train set had 2300 vectors, test set had 687. There was *no median filter*.

The difference between 6.2.1.1.2 and table 6.2.1.1.4 is that the later one didn't use sorting of train set, which caused the rate decreased by 5.53%.

6.2.1.2. The following group is test results of one_hidden_layer neural networks with 40 hidden nodes, **include neutral**, **not take peak**, no momentum number, combine with either sorting the train data or not.

| Emotion | Neutral | Joy | Surprise | Angry | Disgust | Fear | Sad |
|---|---|---|---|---|---|---|---|
| Neutral | 77. 69 | 0. 21 | 2. 48 | 2. 27 | 9. 92 | 2. 27 | 5. 17 |
| Joy | 0. 00 | 79. 51 | 0. 00 | 10. 66 | 0. 00 | 9. 84 | 0. 00 |
| Surprise | 0. 00 | 0. 00 | 92. 93 | 0. 00 | 0. 00 | 0. 00 | 7. 07 |
| Angry | 2. 04 | 1. 36 | 0. 00 | 76. 19 | 19. 05 | 0. 00 | 1. 36 |
| Disgust | 1. 14 | 3. 41 | 1. 14 | 6. 82 | 75. 00 | 1. 14 | 11. 36 |
| Fear | 1. 90 | 28. 57 | 0. 00 | 0. 00 | 4. 76 | 63. 81 | 0. 95 |
| Sad | 30. 16 | 0. 79 | 13. 49 | 3. 97 | 6. 35 | 3. 17 | 42. 06 |
| Average | 72. 45 | | | | | | |

**Table 6.2.1.2.1**

The result is obtained by neural network with 1 hidden layer, 40 hidden nodes, σ = 0.91, α = 0.7 and had *median filter which size was 3*.

Train data set *included the neutral emotion and was sorted which number was 1526*.

Test vectors number is 1171.

Before sorting and selection, train set had 3200 vectors.

| Emotion  | Neutral | Joy   | Surprise | Angry | Disgust | Fear  | Sad   |
|----------|---------|-------|----------|-------|---------|-------|-------|
| Neutral  | 76. 65  | 2. 89 | 1. 03    | 4. 55 | 8. 26   | 1. 03 | 5. 58 |
| Joy      | 5. 74   | 58. 20| 0. 00    | 4. 10 | 10. 66  | 21. 31| 0. 00 |
| Surprise | 0. 00   | 1. 01 | 82. 83   | 0. 00 | 0. 00   | 6. 06 | 10. 10|
| Angry    | 2. 72   | 0. 68 | 0. 00    | 65. 99| 29. 25  | 0. 00 | 1. 36 |
| Disgust  | 13. 64  | 0. 00 | 0. 00    | 13. 64| 65. 91  | 2. 27 | 4. 55 |
| Fear     | 5. 71   | 30. 48| 0. 00    | 0. 95 | 12. 38  | 50. 48| 0. 00 |
| Sad      | 30. 16  | 4. 76 | 3. 97    | 2. 38 | 7. 14   | 0. 00 | 51. 59|
| Average  | 64. 52  |       |          |       |         |       |       |

**Table 6.2.1.2.2**

The result is obtained by neural network with 1 hidden layer, 40 hidden nodes, median *filter size = 3*,

σ = 0.91, ɑ = 0.61.

Training set had 3200 vectors *not sorted, not selected* and *included the neutral emotion*.

The test set had 1171 vectors.

By comparing table 6.2.1.2.1 and table 6.2.1.2.2, all the other parameters are the almost same except the previous one sorted and selected the train set, which led to 7.93% higher rate than the later one. In this case the sorting played an important role for improving the rate.

6.2.1.3 The following group is result of one_hidden_layer neural networks with 40 hidden nodes, **include neutral, take peak frames,** pick frame size was 5 which means we took 5 frames at the peak. There was no momentum number. It combined with either using median filter or not.

| Emotion | Neutral | Joy | Surprise | Angry | Disgust | Fear | Sad |

| Emotion | Neutral | Joy | Surprise | Angry | Disgust | Fear | Sad |
|---|---|---|---|---|---|---|---|
| Neutral | 72. 52 | 0. 00 | 2. 89 | 2. 89 | 13. 43 | 2. 69 | 5. 58 |
| Joy | 1. 64 | 55. 74 | 0. 00 | 8. 20 | 3. 28 | 27. 87 | 3. 28 |
| Surprise | 0. 00 | 0. 00 | 90. 91 | 0. 00 | 0. 00 | 1. 01 | 8. 08 |
| Angry | 5. 44 | 0. 00 | 0. 00 | 51. 70 | 36. 05 | 2. 04 | 4. 76 |
| Disgust | 3. 41 | 3. 41 | 0. 00 | 14. 77 | 57. 95 | 2. 27 | 18. 18 |
| Fear | 11. 43 | 12. 38 | 1. 90 | 0. 95 | 12. 38 | 57. 14 | 3. 81 |
| Sad | 25. 40 | 0. 79 | 16. 67 | 6. 35 | 12. 70 | 2. 38 | 35. 71 |
| average | 60. 24 | | | | | | |

**Table 6.2.1.3.1**

The result is obtained by neural network with no *median filter*, 1 hidden layer, 40 hidden nodes,

σ = 0.81,  α = 0.61.

The test set had 1171 vectors. The train set had 2300 include *the neutral emotion, after taking peak frames the train set number decreased to 1026*, then *after sorting it further decreased to 490.*

| Emotion | Neutral | Joy | Surprise | Angry | Disgust | Fear | Sad |
|---|---|---|---|---|---|---|---|
| Neutral | 74. 38 | 0. 00 | 2. 69 | 2. 48 | 13. 22 | 2. 27 | 4. 96 |
| Joy | 1. 64 | 58. 20 | 0. 00 | 8. 20 | 1. 64 | 27. 05 | 3. 28 |
| Surprise | 0. 00 | 0. 00 | 91. 92 | 0. 00 | 0. 00 | 0. 00 | 8. 08 |
| Angry | 6. 12 | 0. 00 | 0. 00 | 53. 74 | 35. 37 | 0. 68 | 4. 08 |
| Disgust | 3. 41 | 1. 14 | 0. 00 | 13. 64 | 61. 36 | 2. 27 | 18. 18 |
| Fear | 11. 43 | 12. 38 | 2. 86 | 0. 95 | 11. 43 | 58. 10 | 2. 86 |
| Sad | 27. 78 | 0. 79 | 16. 67 | 6. 35 | 12. 70 | 2. 38 | 33. 33 |
| average | 61. 58 | | | | | | |

**Table 6.2.1.3.2**

The result is obtained by neural network with 1 hidden layer, 40 hidden nodes,  σ = 0.81,  α = 0.61,

with *median filter which size was 3*.

The test vectors' number was 1171. The train set had 2300 vectors, *including the neutral emotion, after taking peak frames*, the number decreased to 1026, after *sorting and selecting*, the number of train data was 490.

By comparing table 6.2.3.1 and table 6.2.3.2, it is easy to see that all the other neural network parameters are similar except table 6.2.1.3.2 used median filter. This caused the one using median filter has 1.34% higher rate.

6.2.1.4 The following group is result of one_hidden_layer neural networks with 40 hidden nodes, **exclude neutral, take peak frames**, no momentum number, peak size =5, combine with either using median filter and sorting the train data or not.

| Train set | After take peak frame | After sort | Test set size | Rate (filter size 3) | Rate no filter |
|---|---|---|---|---|---|
| 3200 | No peak frame | 1962 | 687 | 67. 4% | 67. 3% |
| 3200 | 1448 | 972 | 687 | 63. 9% | 63. 2% |
| 2300 | 1030 | No sorting | 687 | 63. 9% | 65. 1 |
| 2300 | No peak frame | 1446 | 687 | 75. 6% | 73. 4 |
| 3200 | 1448 | No sorting | 687 | 66. 6% | 64. 1% |

**Table 6.2.1.4.1**

Test results from 5 different neural nets on Cohn_Kanade database.

From this table, we can see taking peak frames always cause lower rate. And by comparing the first row and the fourth row, we can also see the rate decreased from 73.4% to 67.3% while train number raised from 2300 to 3200. This can be explained as over-training. By comparing column 5 and column 6, we can infer the median filter improved the rate in all cases except in the case of third row.

Generally, the group of tables in chapter 6.2.1.1 is result when **excluding neutral, not taking peak frames of data**. The group of tables in chapter 6.2.1.2 is result when **including neutral**, **not taking peak frames of data**. The group of tables in chapter 6.2.1.3 is result when **including neutral**, **taking peak frames of data.** The group of tables in chapter 6.2.1.4 is result **excluding neutral and taking peak frames of data.** In each group, we also tested the different combination with using median filter or not, sorting the train set or not.

From various compares, we can conclude that taking the peak frames will lead to lower rate, while median filter, sorting of train set and excluding neutral will improve rate. From table 6.2.1.4.1, we can further discover that sorting and not taking peak frames can significantly affect the results, normally the contribute was more than 2%. On the other hand, from the compare between group 6.2.1(not taking peak, no neutral) and group 6.2.2 (not taking peak but with neutral), we can infer that excluding neutral emotion or not doesn't make too much difference.

We can easily find what need to be compared by choosing different pairs of tables, thus find interesting proof to our design ideas in chapter 5. Readers can also verify our conclusions by freely choosing proper pair from the numerous tables to compare.

### 6.2.2 Neural networks for special emotion categories

We infer from the previous results that the emotion surprise always has higher recognition rate than others, one may ask if it is true that this emotion dominates the neural networks' weights thus constraints other emotions' recognition rate? How about try different classifying categories to get clearer understanding to the relationship between the data?

We tried ignoring all the surprise emotion data for its highest recognition rate among others, also ignoring neutral emotion data for its overwhelming amount of data. If the rate of joy, which normally has the second highest rate can improve much higher, then we can do this recursively: one neural network for one emotion, the idea and steps as following:

First detect if it is surprise, if it is, then stop, otherwise, try the second neural network, to classify if it is joy, if it is, then stop, otherwise repeat the previous steps until it matches. If the recognition rate for all the emotions is higher than 90% just as we has for surprise, then the average rate might be more satisfactory, which means the assumption that emotion with higher rate restraints others' rate is correct.

After ignoring surprise and neutral, the rate of joy rises, but not too much. In the cases we tested, the rate for joy improved from around 75% to 81.4%--84%, thus we stopped at this method.

Second, we tried different category of the emotions to infer more about the relationship between different emotion categories, see following tables. From those tables, we didn't find higher classifying rate by reset the category of these emotions, which means if we design 7 neural networks for each emotion will not help in improving the accuracy, proof can be found from table 6.2.2.2 to table 6.2.2.7.

|          | Neutral | Positive | Surprise | Negative |
|----------|---------|----------|----------|----------|
| Neutral  | 79.05%  | 3.32%    | 0.83%    | 16.80%   |
| Positive | 13.11%  | 76.23%   | 0.00%    | 10.66%   |
| Surprise | 4.04%   | 0.00%    | 92.93%   | 3.03%    |
| Negative | 11.16%  | 10.73%   | 1.29%    | 76.82%   |

**Table 6.2.2.1**

Result on Cohn-Kanade data set.

Network parameter: 1 hidden layer, 10 hidden nodes, $\lambda = 0.5$, $\alpha = 0.5$, $\sigma = 0.3$.

Test vector number was 687, train vector number was 3756, use median filter and normalize output.

Table 6.2.2.1 is constructed to see how much the neural network can differentiate between these 4 categories, where positive emotion means joy, negative emotion includes disgust, anger and fear. The average rate is higher than classifying them into 7 categories.

|         | Joy    | Not joy |
|---------|--------|---------|
| Joy     | 73.77% | 26.33%  |
| Not joy | 4.39%  | 95.61%  |

**Table 6.2.2.2**

Result on Cohn-Kanade data set.

Network parameter:1 layer, 10 hidden nodes, $\lambda = 0.55$, $\alpha = 0.6$, $\sigma = 0.3$.

Test vector number 687, train vector number 3756, use median filter and normalize output.

|          | Surprise | Not surprise |
|----------|----------|--------------|
| Surprise | 87.82%   | 12.12%       |
| Not surprise | 0. 93% | 99. 07% |

**Table 6.2.2.3**

Result on Cohn-Kanade database.

Network parameter: 1 hidden layer, 10 hidden nodes $\lambda = 0.55$, $\alpha = 0.6$, $\sigma = 0.3$.

Test vector number 687, train vector number 3756, use median filter and normalize output.

| | Angry | Not angry |
|---|---|---|
| Angry | 48. 3% | 51. 7% |
| Not angry | 2. 05% | 97. 95% |

**Table 6.2.2.4**

Result on Cohn-Kanade database.

Network parameter: 1 hidden layer, 10 hidden nodes, $\lambda = 0.55$, $\alpha = 0.6$, $\sigma = 0.3$.

Test vector number 687, train vector number 3756, use median filter and normalize output.

| | Disgust | Not disgust |
|---|---|---|
| Disgust | 55. 68% | 44. 32% |
| Not disgust | 5. 08% | 94. 92% |

**Table 6.2.2.5**

Result on Cohn-Kanade database.

Network parameter: 1 layer, 10 hidden nodes, $\lambda = 0.55$, $\alpha = 0.6$, $\sigma = 0.3$.

Test vector number 687, train vector number 3756, use median filter and normalize output.

| | Fear | Not fear |
|---|---|---|
| Fear | 52. 38% | 47. 62% |
| Not fear | 1. 78% | 98. 22 |

**Table 6.2.2.6**

Result on Cohn-Kanade database.

|         | Sad     | Not sad |
|---------|---------|---------|
| Sad     | 32. 71% | 66. 29% |
| Not sad | 2. 39%  | 97. 61% |

**Table 6.2.2.7**

Result on Cohn-Kanade database.

Network parameter: 1 hidden layer, 10 hidden nodes, λ = 0.55, α =0.6, σ =0.3.

Test vector number 687, train vector number 3756, use median filter and normalize output.

## 6.3   Results of test on Cohn-Kanade database ---different nodes and layers

The following tables are results of test on various neural networks with different nodes, layers. The performance of networks varies with its different size. Except for comparison purpose, we try to avoid very large number of hidden nodes, which cause lower generalization of network.

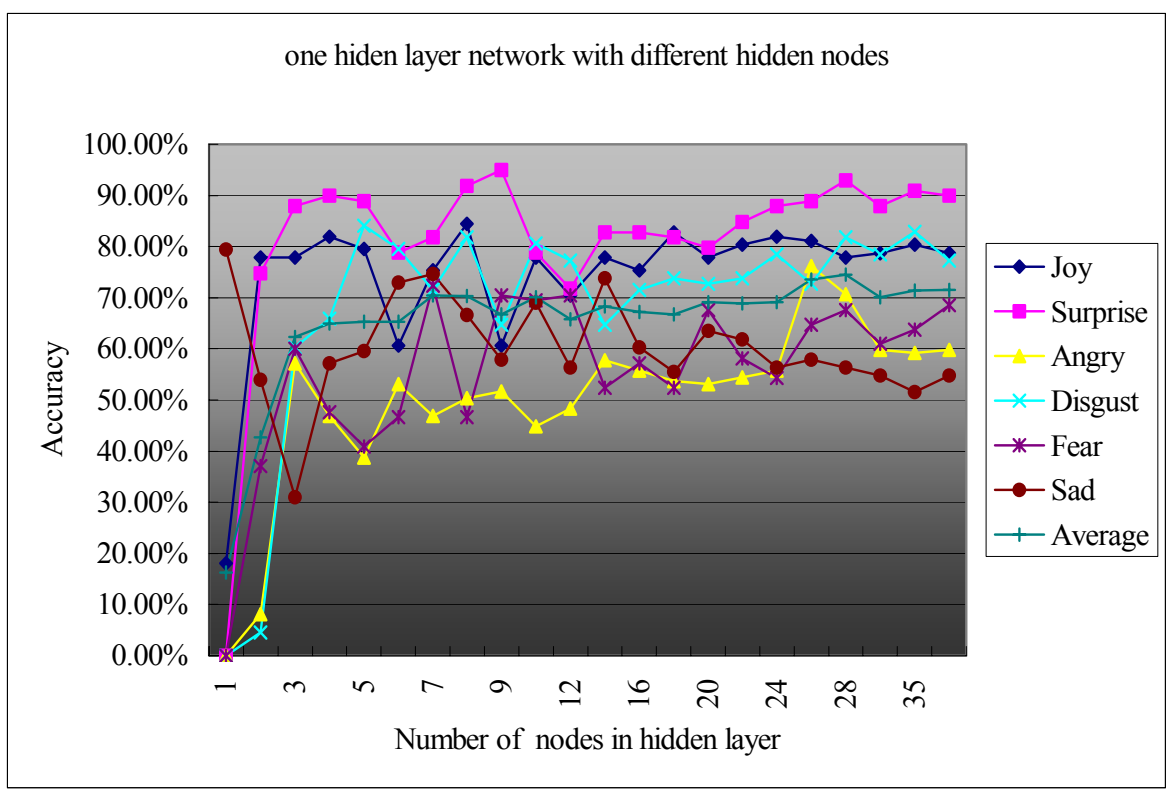

**Figure 6.3.1**

Test on one_hidden_layer network with different nodes in hidden layer

In this figure, number of test vectors was 687; number of train vectors was 3756. It used median filter and normalized output, but no momentum number. Each rate was obtained by values of α and σ. The value of α and σ in this figure was not recorded by code. On the X-axis, the hidden nodes interval before 10 hidden nodes is 1, between 10 and 20 is 2 nodes per grid on the X-axis, and 5 nodes after 20 hidden nodes.

We can infer from the figure, the average classifying rate fluctuates while the number of node changes. This proves that the number of nodes is another important factor to neural network. We may also notice that the neural network with 40 nodes now obtains the best accuracy in this figure, but this is not always true while changing the value of α and σ.

One interesting thing we noticed is that although some papers reported 2_hidden_layer neural nets are normally better than 1_hidden_layer neural nets [14], while selecting different learning rate, momentum number, activation function and its parameter, during our experiments, it is hard to say which one outperforms another. See figure 6.3.2 and table 6.3.1. Comparing with the rate obtained from 1_hidden_layer, we failed to report a better rate by 2_hidden_layer neural nets. The reason is that the automatic selection of neural network parameters worked better in 1_hidden_layer than in 2_hidden_layer, since there is one more parameter – the second hidden layer' nodes number which caused the parameter combination more complex thus more time consuming to train more neural nets. It is most likely that we can find better model by trying more combinations of 2_hidden_layer.

From table 6.3.1, one assumption might be drawn that there is data inconsistency between train set and test set in Cohn_Kanade database, since the train rate can be achieved by more than 98%, but the test rate is only more than 60%. This can be explained as over fitting of the neural nets, but the true reason for over fitting is the data inconsistency. The better consistency between train and test set, the better back propagation works. If there is an effective way to solve the inconsistency, the emotion recognition rate can be improved much more.

Another conclusion we can draw from table 6.3.1 is that our 2_hidden_layer networks works properly, since they can achieve near 100% train rate, which means the errors being decreased almost to zero by proper propagation of their differential.

| Nodes (hidden1Xhidden2) | Train rate | Test rate | Train vector number | Test vector number | σ range | σ increase Step | α range | α increase Step | λ range | λ increase step |
|---|---|---|---|---|---|---|---|---|---|---|
| 29x28 | 94.6 | 70.2 | 3756 | 687 | 2-2.5 | 0.5 | 0.2-1 | 0.3 | 0.5-1 | 0.3 |
| 29x29 | 96.9 | 68.5 | 3756 | 687 | 2-2.5 | 0.5 | 0.2-1 | 0.3 | 0.5-1 | 0.3 |
| 29x29 | 97.7 | 66.3 | 3756 | 687 | 1.8-2.5 | 0.5 | 0.2-1 | 0.3 | 0.5-1 | 0.3 |
| 4 x 4 | 81.83 | 67.32 | 3756 | 687 | 0.9-2 | 0.5 | 0.2-1 | 0.3 | 0.5-1 | 0.3 |
| 4 x 5 | 82.77 | 69.7 | 3756 | 687 | 0.9-2 | 0.5 | 0.2-1 | 0.3 | 0.5-1 | 0.3 |
| 29x29 | 98.7 | 66.5 | 3756 | 687 | 0.9-2 | 0.3 | 0.2-1 | 0.3 | 0.5-1 | 0.3 |

**Table 6.3.1**

Test result on 2_hidden_layer: different nodes and different activation parameter categories

Neural net parameters: test vector number 687, train vector number 3756.

Use median filter but no normalize output, each rate is with different set of $\lambda$, $\alpha$ and $\sigma$.

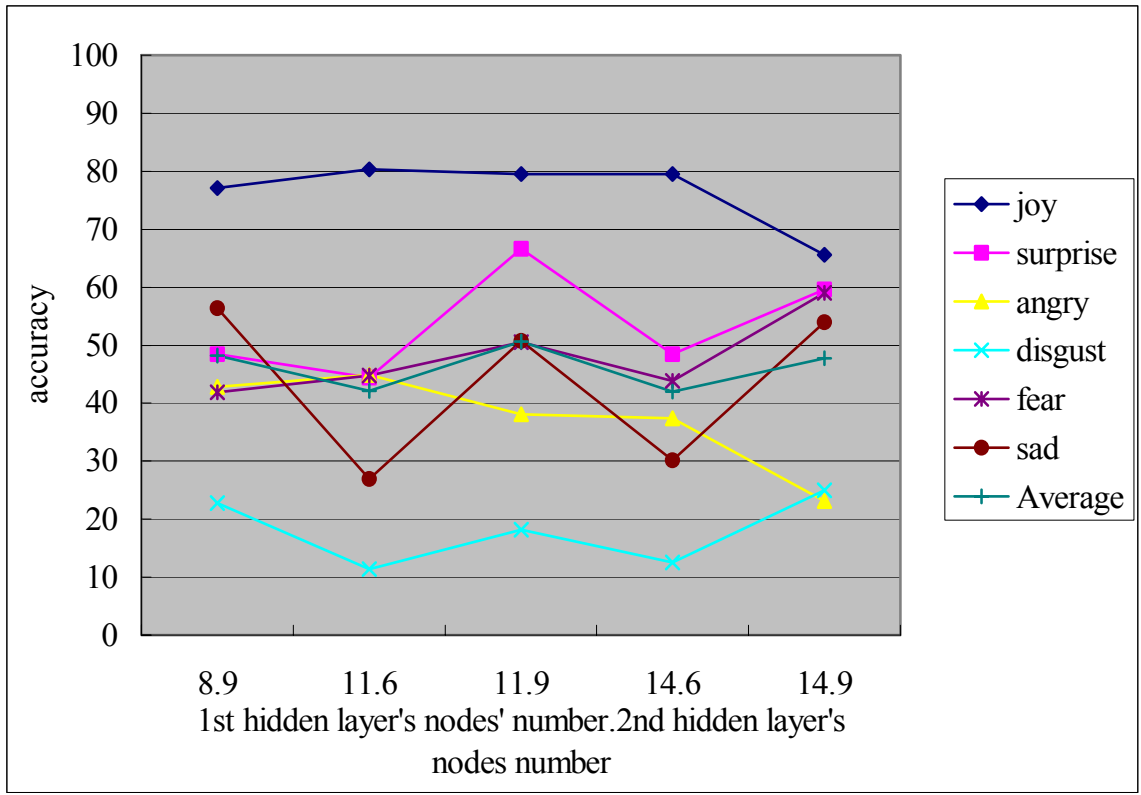

**Figure 6.3.2**

Five 2_hidden_layer neural networks with improper weights threshold [0.5, -0.5]

not only for initialization, but also restrain the weights to this threshold all through the training procedure.

Activation function parameter category [0.1-1] is not proper for this case, thus caused neuron saturate, and led to low recognition rate. Since it is not a successful result, the detailed combination of parameters was not kept.

|  | Joy | Surprise | Angry | Disgust | Fear | Sad |
|---|---|---|---|---|---|---|
| Joy | 63.11 | 5.74 | 7.38 | 4.10 | 19.67 | 0.00 |
| Surprise | 0.00 | 96.97 | 0.00 | 0.00 | 2.02 | 1.01 |
| Angry | 2.04 | 0.00 | 79.59 | 16.33 | 0.00 | 2.04 |
| Disgust | 3.41 | 5.68 | 21.59 | 65.91 | 2.27 | 1.14 |
| Fear | 25.71 | 0.00 | 0.95 | 0.95 | 71.43 | 0.95 |
| Sad | 0.00 | 3.17 | 3.97 | 8.73 | 0.00 | 84.13 |
| average | 76.86 |  |  |  |  |  |

**Table 6.3.2**

Best model: 1 hidden layer, 4 hidden nodes, α =0.3, λ =0.6, σ =0.9, test on Cohn-Kanade database.

Number of train data is 2500, after sorting, number of train data decreased to 1476.

Number of test database is 687, with median filter (size =3), not took peak frames of train and test data.

Here comes the best set of parameters in table 6.3.2, with highest rate among other neural nets in this thesis, also with small learning rate which means the generalization could be fine if change test database, the momentum number is normal comparing with the value 0.5 reported by others. Hence proved that a good set of parameters is very important in design neural networks. We also tried change the activation function parameter from 0.9 to 1 which is a normal value by other reports, the rate will decrease by around 2%, this proved our way to select the combination of parameters works better than just empirically selecting some combination.

## 6.4 Results of test on Cohn-Kanade data set ---different initial weights

As mentioned in chapter 5.2, different starting point of the weights might cause different results, but how much will the weights' initialization affect the neural networks? The fluctuations of the rate curve in the following tables are part of the answers, where the value n of "initialization time" means this is the nth time we reset the starting points.

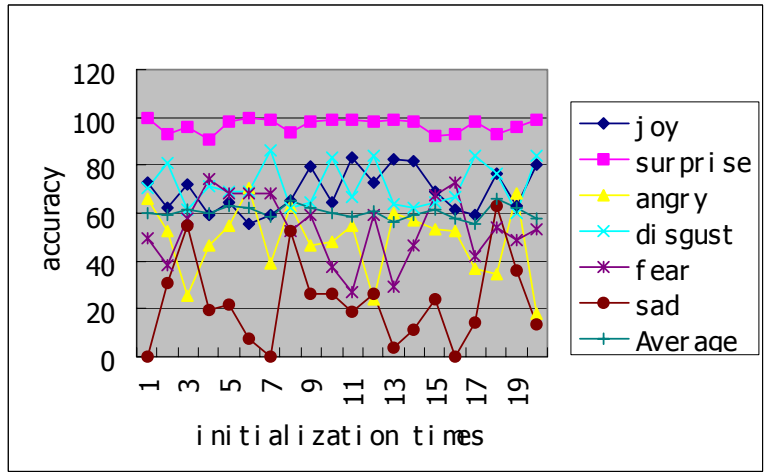

**Figure 6.4.1**

1 hidden layer with 10 hidden nodes, α =0.4, λ = σ =0.6, initialize weights 20 times.

Normalize output and use median filter.

We can also see the rate curve in figure 6.4.1 fluctuates more than that in figure 6.4.2 where normalization of output is not used in the later one. But sometimes the rate in table 6.4.1 is a little bit higher which means it is easier to find higher rate by normalization.

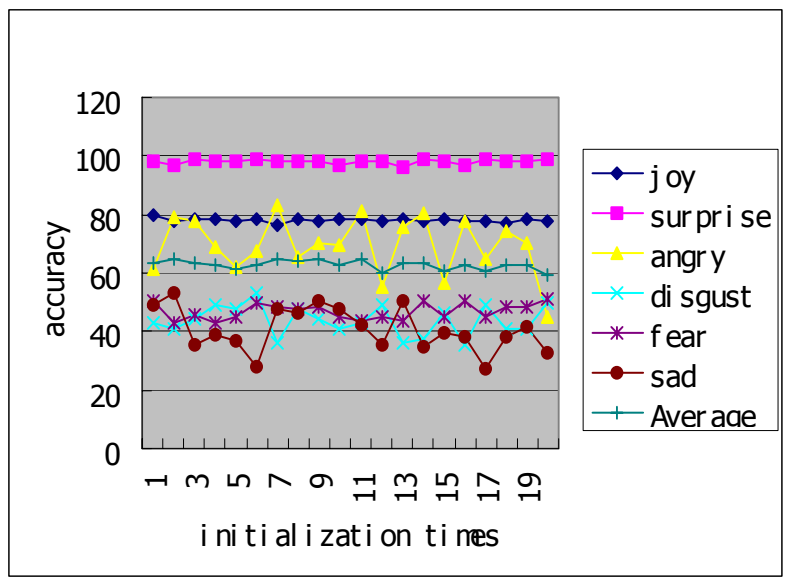

**Figure 6.4.2**

1 hidden layer with 10 hidden nodes, α = 0.4, λ = σ =0.6, initialize weights 20 times,

not normalize output, using median filter.

# 7  Discussion

This thesis presents back propagation neural networks for emotion classifying. We tried quite a few ways, but there still exists further improvement to this work, i.e. add a better interface for manually controlling the parameters during training.

There are many different factors, which might affect the performance of a neural network. Even for those factors we mentioned in this thesis, we didn't exhaust all these combinations, such as:

1) The median filter can consider more neighboring frames.
2) Design different neural networks for features of upper face and lower face respectively, and combine their results.
3) Use simulated annealing instead of Powell's direction set to minimize the error function.
4)  Handle over-training better.
5) Change the activation function from sigmoid to others.

As for the authentic data set, if there are 7 emotions, it is clearer to compare with the results we get from Cohn-Kanade's data set. Though some of the natural emotions are very difficult to get, i.e. to make people angry without telling them it is only an experiment.

Nowadays, there are lots of other people doing similar work using back propagation. The recognition rate of their systems averages from 67% to 100% [4,14]. But it is unfair to compare the results between different system on different database and different constrains in experiments while obtaining the database. From the paper we surveyed, none of these neural networks did emotion classifying on Cohn-Kanade data set or on authentic database.

Another interesting thing is that we didn't find any report about using neural networks analyzing Cohn-Kanade data by holistic ways instead of analytic ways, which we used here. While it is referred that holistic ways outperforms the later one [30,31,32]. It will be interesting if we use the whole face's information instead of the analytic facial features as the input of the neural network.

Most of the automatic emotion analysis system classifies human emotions by 7 prototypic emotions[13].  One main drawback of this method is that how to handle the blended emotions or subtle facial changes, i.e. surprise with fear or surprise with joy. A feasible solution might be using more categories--- more than 7 output nodes in neural nets. Since not trying this way, we have no idea how to set up the categories. This need psychology research results..

Recently, there are reports focusing on using facial component models to model and track the fine-grained changes in facial expression [13]. The recognition rate for the facial action unit (AU) is 93% - 96% on Cohn-Kanade data set [13]. But there are more than 7000 combinations of Aus[14], thus even if the AUs are correctly recognized, how to let computer infer from these 7000 combinations what people feel will be a step before computer finally communicate with human by facial expressions. However the accurate recognition of AUs is a very important progress. Classifying emotions from the AU combinations instead of from the image or the features directly might lead to more precise emotion recognition in the future. Another future direction is to use salient points [33] to aid in the tracking precision. Together with other detection methods for heart beat, voice, body gesture recognition, contents of communication, there will be a day for people to enjoy interaction with a fully humanism computer.